УДК 004.83

**Artem Kramov**

Taras Shevchenko National University of Kyiv

Academika Hlushkova Ave, 4G, 03022 Kyiv, Ukraine

## Evaluating text coherence based on the graph of the consistency of phrases to identify symptoms of schizophrenia

Different state-of-the-art methods of the detection of schizophrenia symptoms based on the estimation of text coherence have been analyzed. The analysis of a text at the level of phrases has been suggested. The method based on the graph of the consistency of phrases has been proposed to evaluate the semantic coherence and the cohesion of a text. The semantic coherence, cohesion, and other linguistic features (lexical diversity, lexical density) have been taken into account to form feature vectors for the training of a model-classifier. The training of the classifier has been performed on the set of English-language interviews. According to the retrieved results, the impact of each feature on the output of the model has been analyzed. The results obtained can indicate that the proposed method based on the graph of the consistency of phrases may be used in the different tasks of the detection of mental illness.

### Introduction

Natural language processing (NLP) problems can be seen in different areas of professional activity. Means of processing heterogeneous textual information are utilized to evaluate text quality, to analyze customers' reviews, to search necessary data, etc. Due to the availability of the ambiguity elements of a text and the lack of its unified structure (any word order, style of speaking) certain tasks of NLP cannot be solved with the usage of defined algorithms. The solution of these tasks implies the detection of necessary regularities between text elements, which allow the retrieving of an output result according to a required error. The search for these regularities and text properties is performed using the machine learning methodology. In order to solve corresponding tasks, the training dataset is previously prepared with the further learning of a designed model. Tasks of this type incorporate *the estimation of text coherence*.

The term "coherence of a text" implies its thematic integrity: the communicative ability of an author to convey the main idea of the text to a reader. Apart from the semantic consistency of text elements, the term "coherence" interprets the consistent perception of information by a reader according to existing background knowledge. Another criterion of the achievement of text coherence is the availability of structural connectivity of a text (its *cohesion*). The clear order of elements (words, phrases) allows simplifying the perception of information. The conformity of a text with the mentioned criteria can denote the ability of an author to concentrate around a single topic and to formulate statements. Automated methods of the evaluation of text coherence are utilized as the additional instrument to analyze the properties of a text in different areas. The availability of the error of methods determines the usage of these methods just for previous analysis of text information that helps a specialist to work with an input dataset. The example of such a usage of the text coherence model is the process of detection of schizophrenia symptoms.

The detection of symptoms of mental diseases and the classification of its type are complex processes that require the corresponding qualification of a specialist to solve (for instance, almost 30% of patients [1] with bipolar disorder and schizophrenia are misdiagnosed). The analysis of patients' speech is a part of a complex process of the diagnosis of schizophrenia symptoms: alogia (poverty of speech), lack of permanent focus on a topic (tangential speech), incoherence language, regular usage of metaphors, etc. The usage of NLP methods allows calculating corresponding metrics and applying the methods of machine learning for the prediction of the degree of disease. However, the usage of pre-trained models cannot be considered as an effective solution now due to the low number of samples to train a model. The accuracy of methods is calculated on the training dataset using leave-one-out cross-validation. Such an approach degrades the ability of a model to analyze samples not seen before. Thus, the actual task implies the search for text properties – metrics that can divide the speech of health people from ill patients. Further, these metrics can be used to generate a training dataset to create an effective model of diagnostics of schizophrenia symptoms.

**The purpose of the paper** is the following:

- to perform the comparative analysis of state-of-the-art methods of diagnostics of schizophrenia based on the evaluation of text coherence and other linguistic features;
- to apply the graph of the consistency of phrases to take into account the impact of cohesion and coherence of a text on the effectiveness of texts classification into 2 categories: healthy people and ill persons;
- to perform the comparison of the impact of retrieved metrics with other text properties.

**Schizophrenia detection methods based on the analysis of the coherence of a text**

The principle of work of state-of-the-art methods is based on the following models: tangentiality model [2] and incoherence model [3]. While using the *tangentiality model*, the analysis of the level of the conformity of answers with questions is performed. The text of an answer is divided into "windows of words" – phrases with a fixed count of words. The formalized representation of the question and "windows" is implemented with the usage of Latent Semantic Analysis (LSA): words are represented as the vectors of a semantic space; the formalization of phrases is performed by calculating the average value of corresponding word vectors. The similarity measure between each "window" and the question is estimated as a cosine distance between vectors. Using retrieved distances the linear regression model is built. Then a line slope is analyzed: steeper slope means that the topic of the response is moving away from the question. Such a deviation can indicate the inability of a patient to focus his thoughts (flight of ideas).

The *incoherence model* performs the analysis only of the answer of a patient. A text is divided into the set of sentences. The vector representation of sentences is performed with the usage of LSA in the same manner as in the tangentiality model. The similarity measure of sentences is interpreted as a cosine distance of corresponding vectors. The coherence of an answer is calculated as a minimum similarity measure between adjacent sentences within the text (so-called *first-order coherence*, FOC). Moreover, in order to train and test a classification model (convex hull classifier) other features are utilized: maximum length of a sentence and weighted frequency of the usage of determiners. Both the incoherence model and tangentiality model should be considered as the models of semantic coherence, therefore, their accuracy depends on the semantic representation of the text's elements. In the paper [4], the experimental verification of the effectiveness of different semantic embedding models was checked on the set of English transcripts. The impact of a semantic embedding model was examined for both the tangentiality model and the incoherence model. According to the results [4], the highest accuracy was achieved with the usage of Smooth Inverse Frequency (SIF). Moreover, it is suggested to perform the additional analysis of coreferent objects to detect ambiguous pronouns:

- words that refer to an object that is not mentioned in a text;
- pronouns that refer to an object that is mentioned in a text later (cataphora).

The availability of ambiguous pronouns is not the explicit mark of schizophrenia symptoms but it can be taken into account as an additional characteristic. In the papers [5] and [6], the verification of both models has been performed for German and Russian corpora correspondingly. It should be mentioned that the effectiveness of approaches for German and Russian corpora is different. For example, the incoherence model with the usage of the TF-IDF (semantic embedding model – GloVe) shows the highest accuracy for German texts. In contrast, the tangentiality model was the most successful for Russian texts. Thus, the effectiveness of considered models depends on the features of a certain language (impact of word order, capital letters, etc.) and the accuracy of the semantic embedding model.

Apart from the analysis of semantic coherence, in the paper [7] it is suggested to analyze other linguistic properties of a text to investigate their correlation with estimations of the Social Skills Performance Assessment task. Such properties describe the following features: lexical diversity, lexical density, and syntactic complexity. Mentioned properties allowed to build a classification model to distinguish interviews of healthy people and ill patients. Moreover, it turned out that taking into account these linguistic features increase the accuracy of the distinction of schizophrenia symptoms and bipolar disorder. In the paper [8], it is proposed to consider the impact of metaphors, namely, to take into account the frequency of their usage. The frequent usage of phraseologies or idioms can indicate a mental disease. In spite of the low ratio of metaphors to the general count of words (6.3%), this feature of a text can be considered as an additional characteristic during the training process of a classifier.

After the analysis of different methods of the identification of schizophrenia symptoms based on the estimation of text coherence, the following conclusions can be drawn:

1. The coherence of a text is considered as semantic integrity of the text's elements. The effectiveness of the tangentiality model and the incoherence model depends on the choice of semantic embedding model and the unique features of the language of the text.

2. In order to take into account the repeats of phrases, certain semantic embedding models are used considering the frequency of words mentions. However, such an approach requires the usage of an additional textual corpus to retrieve necessary statistic data.

3. In spite of the availability of other linguistic features of a text (lexical diversity, lexical density, the frequency of metaphors' mentions), current methods do not take into account the structural consistency of text's elements – cohesion. For instance, the lexical density is calculated as the ratio of the count of informative words (nouns, verbs, adverbs) to its general count. Thus, the order of words within a text and the availability of necessary additional elements (determiners, connectors) are neglected.

In the next section, the method of the estimation of the text's coherence based on the graph of the consistency of phrases is suggested to detect the schizophrenia symptoms.

## Detection of schizophrenia methods based on the graph of consistency of phrases

In contrast to previously considered methods that calculate the similarity measure of sentences using the analysis of the similarity of corresponding informative words, it is suggested to analyze the semantic and structural consistency of the text's elements at the level of phrases. Actually, phrases interpret the informative part of a text that allows taking into account the significance of words notwithstanding the part of speech. Moreover, the extraction process of phrases from a sentence depends on the correct order of elements, that is the verification of the structural consistency of words is additionally implemented. Let us consider the process of phrases extraction in more detail.

## Extraction of phrases from a sentence

Firstly, represent a sentence $S$ as the set of tokens (words):

$$S = \{t_1, t_2, ..., t_N\} \tag{1}$$

where $N$ is the count of the sentence's words. The task consists in the representation of the sentence $S$ as the set of phrases where each phrase is the set of tokens: $\{t \mid t \in S\}$. The word order of the sentence is not fixed therefore the applying of defined templates (regular expressions) cannot be an effective solution. Such an approach may be used as the additional instrument to search for phrases according to the style or unique features of speech. In order to detect phrases in English texts, it is proposed to utilize a method called *open information extraction* (open IE). It extracts the phrases of a text by analysis of the syntactic tree of a sentence [9]. The output of the method is the set of tuples where each tuple consists of the following elements: "object" (main element), "relation" (tokens that represent the dependence on the main element), and "subject" (subordinate element).

$$C = \{C_1, C_2, ..., C_K\}$$
$$C_i = (Obj, Sub, Rel), i \in I = \{1, 2, ..., K\} \tag{2}$$

where $K$ is the count of extracted tuples. After the union of the elements of the tuple $C_i$ the following phrase can be retrieved: $X_i = Obj \cup Rel \cup Sub$. Thus, the sentence $S$ can be represented as the set of phrases:

$$X = \{X_1, X_2, ..., X_K\} \tag{3}$$

For example, a sentence "He sits with Kyle while he eats." can be represented by the following phrases:

$$X = \{X_1, X_2\}$$
$$X_1 = \{\text{He, sits, with, Kyle}\} \tag{4}$$
$$X_2 = \{\text{He, sits, he, eats}\}$$

## The graph of the consistency of phrases

Let us consider sentences $S_i$ and $S_j$ that are represented by the sets of phrases $X^i$ and $X^j$ correspondingly. The permanent repeats of phrases are the one of schizophrenia symptoms. In the case of the detection of such repeats, it is necessary to decrease the output estimation of the text's coherence. It is suggested to represent the sentence $S_j$ as the relative complement of $X^i$ in $X^j$: $X^j \setminus X^i$. In such a case, all phrases of the sentence $S_j$ that present in the sentence $S_i$ are removed. Such an approach allows taking into account the negative impact of repeats in adjacent sentences due to the decrease of the informative part of a text. Moreover, it is necessary to detect the impact of repeats within a single sentence. Let us formulate sets $A$ and $B$ that consist of the unique phrases of each sentence:

$$A = \text{unique}(X^i)$$
$$B = \text{unique}(X^j \setminus X^i)$$

(5)

In addition, it is necessary to define a function count that returns the number of times the phrase is repeated within a sentence. Then a complete directed bipartite graph $K_{ij}$ is build from the sets of phrases $A$ and $B$. Figure 1 shows the example of the graph $K_{ij}$.

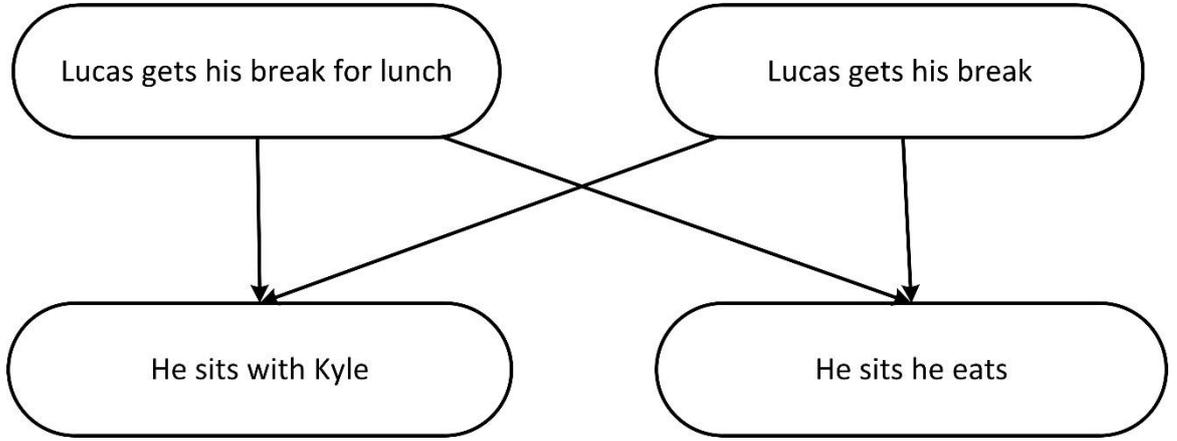

Рис. 1 – The example of the graph of the consistency of phrases for $S_i$ and $S_j$

In order to distinguish the impacts of semantic (semantic coherence) and cohesion parts, two separate graphs of consistency of phrases are built: $K_{ij}^{sem}$ and $K_{ij}^{coh}$. The weight of edges of the graph $K_{ij}^{sem}$ is calculated as a cosine distance between the corresponding vectors of phrases:

$$\text{weight}_{sem}(A_l, B_m) = \frac{\mathbf{A}_l \cdot \mathbf{B}_m}{\|\mathbf{A}_l\| \cdot \|\mathbf{B}_m\| \cdot \text{count}(\mathbf{A}_l) \cdot \text{count}(\mathbf{B}_m)}$$

(6)

where $l \in \left\{1, 2, ..., |A|\right\}, m \in \left\{1, 2, ..., |B|\right\}, l \neq m$. The denominator contains values that interpret the number of times the phrases are repeated within the sentence. In this way, the negative impact of repeats on the estimation of coherence is taken into account. The weight of edges of the graph $K_{ij}^{coh}$ is calculate in the following way:

$$\text{weight}_{coh}(A_l, B_m) = \frac{\text{common}(A_l, B_m)}{\text{unique}(A_l, B_m) \cdot \text{count}(A_l) \cdot \text{count}(B_m)}$$

(7)

where $\text{common}(A_l, B_m)$ is the count of common terms, $\text{unique}(A_l, B_m)$ is the count of unique terms of both phrases. In the case of the availability of coreferent relation [10] between phrases, the value of $\text{common}(A_l, B_m)$ equals to $\text{unique}(A_l, B_m)$. Such an approach allows increasing the significance of coreferent relation within a text. Moreover, in this way, the deviation of speech from the general topic of a text can be tracked.

The similarity measure of the sentences $S_i$ and $S_j$ is described by two values: semantic coherence and cohesion. The mentioned values $\text{sem}(S_i, S_j)$ and $\text{coh}(S_i, S_j)$ are calculate as the average outdegree value of corresponding graphs:

$$\text{sem}(S_i, S_j) = \text{avg}(\text{outdegree}(K_{ij}^{sem}))$$
$$\text{coh}(S_i, S_j) = \text{avg}(\text{outdegree}(K_{ij}^{coh}))$$

(8)

### Estimation of the output metrics of a text

After the calculation of the similarity measure of sentences, it is necessary to estimate metrics that interpret some features of a text. It is proposed to estimate the semantic coherence with the usage of the incoherence model: the output value of the model is the minimum value of the semantic similarity of adjacent sentences within a text − first-order coherence, $FOC$. However, in the paper [11] the necessity of taking into account the connection between all sentences is mentioned. Such an approach allows increasing the accuracy of the method. Thus, let us build a complete directed graph $G = (V, E)$, where $V$ is the set of vertices that interpret the sentences of a text, $E$ is the set of edges. The weight of the edge $e_{ij} \in E, i \neq j$, that connects vertices $i$ and $j$, is calculated in the following way:

$$\text{weight}(e_{ij}) = \frac{\text{sem}(S_i, S_j)}{|i - j|}$$

(9)

The denominator $|i - j|$ is used to normalize edge values to take into account the distance between sentences. The estimation of the semantic coherence of a text $SemCoh$ is calculated as the average value of edge weights:

$$SemCoh = \frac{\sum\limits_{i, j \in \{1, 2, \ldots, |V|\}, i \neq j} \text{weight}(e_{ij})}{|V|}$$

(10)

The estimation of the cohesion part $Cohesion$ is performed the same as the metric $SemCoh$. The only difference consists in the calculation of the edge weights $e_{ij}$ of the graph $G$:

$$\text{weight}(e_{ij}) = \frac{\text{coh}(S_i, S_j)}{|i - j|}$$

(11)

Thus, the following metrics are retrieved with the usage of the analysis of integrity and connectivity: *FOC*, *SemCoh*, *Cohesion*. Moreover, it is worth considering other linguistic features of a text that can indicate the poverty of speech – one of the schizophrenia symptoms. These features include the following metrics that were used to classify texts in the paper [7]:

1. Lexical density *FuncW* that shows the ratio of the number of informative tokens (nouns, pronouns, verbs, etc.) to the general count of words. However, additional elements (prepositions, determiners) are necessary to form and connect phrases. Thus, it is suggested to consider a metric *PhraseW* that is calculated as the ratio of the unique words of phrases to the general count of tokens.

2. Lexical diversity that can be described by the following metrics: *MATTR* (moving average type-to-token ratio) and *BI* (Brunet's Index).

To sum up, seven metrics are analyzed for each text. These metrics form a feature vector for each document. In the next section, the significance of each metric is considered during the training process of a model-classifier.

## Results

The English-language samples of interviews of patients were retrieved from papers [12] and [13]; interviews of healthy people were obtained from the web-resource [14]. A feature vector was formed for each text, a decision tree was chosen as a model-classifier. Parameters of the model were chosen using a grid search. Such a choice can be explained by the ability of this model to perform further analysis of the impact of each feature on the output result. Moreover, the decision tree can be effectively used to analyze a small dataset (100 texts).

In spite of the effectiveness of the combination of models (SIF and Word2Vec, TF-IDF and GloVe), such an approach requires the usage of additional text corpus to retrieve statistical data about the elements of a text. Thus, in order to perform the vector representation of the text's elements, a single pre-trained Word2Vec model was chosen. The chosen model was pre-trained on the text corpus "Google News". The extraction of phrases and the search for coreferent objects in a text was implemented with the usage of the Stanford CoreNLP tools, namely, offline web-server with the built-in API (Application Programming Interface). In order to perform the generation of feature vectors and to train the model-classifier, a Python-language (Python 3.6) application was created.

Figure 2 shows the impact of each feature on the output after the training process. The highest value was retrieved for the feature *SemCoh*. It can indicate the necessity to take into account the connection between all text's sentences (in contrast to *FOC*). The diagram demonstrates the feasibility of the usage of the proposed graph of phrases' consistency because the first two metrics (*SemCoh* and *Cohesion*) were calculated using this graph. The significance of the metric *Cohesion* confirms the necessity to take into account the coreferent relation as the element of the global connectivity of a text. The lack of the mentioned connection and the availability of cataphora can indicate gradual or sharp deviations of the text's content from its general topic. The significance of linguistic text features (*FuncW*, *PhraseW*, *MATTR*, *BI*) does not exceed 0.1 which shows the ability to use these metrics just as an additional indicator.

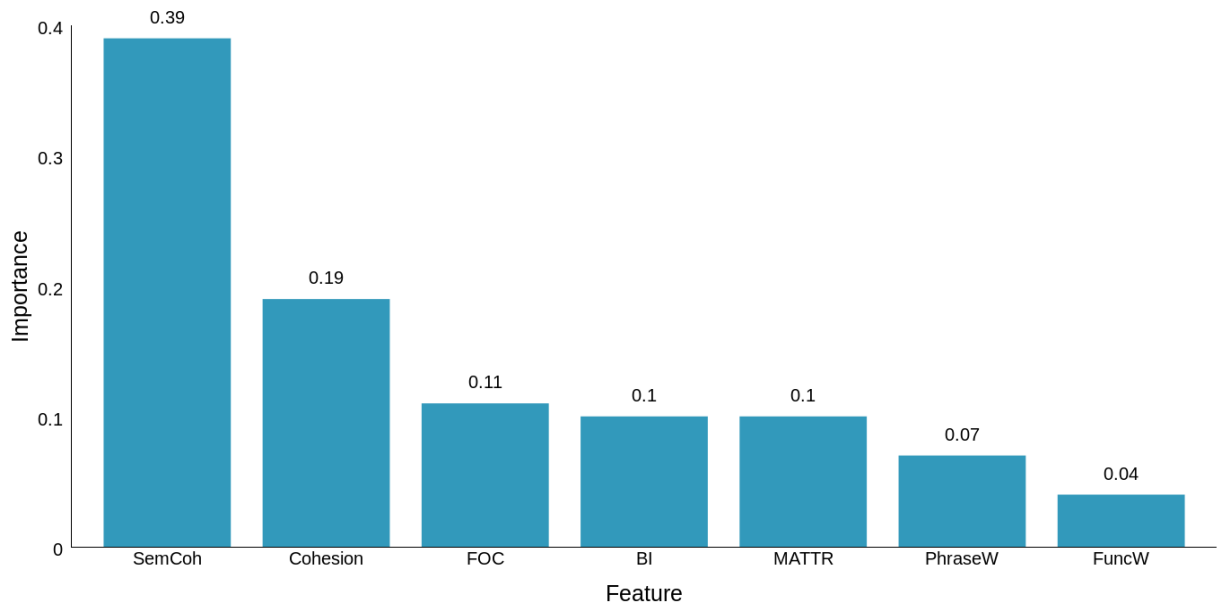

Рис. 2 – The significance diagram of the each feature for the model-classifier

## Conclusions

Taking into account the performed comparative analysis of the methods of the detection of schizophrenia symptoms based on the estimation of text's coherence, and the retrieved results of the impact of different metrics, the following conclusions can be drawn:

- The methods of the detection of schizophrenia symptoms perform the estimation of the text's coherence by the analysis of its semantic integrity. The formalization of the text's elements is implemented with the usage of the combination of different semantic embedding models. Such an approach allows taking into account the availability of repeats of phrases that can identify people with mental illness. However, it requires the availability of an additional textual corpus in order to calculate statistical data.

- Analysis of texts at the level of phrases allows avoiding additional filtering of stop words and taking into account necessary auxiliary elements (e.g. determiners). Moreover, the automated extraction of phrases requires the availability of structural consistency of a sentence that can have implications for an output value of the text's coherence.

- The retrieved values of the significance of metrics based on the graph of consistency of phrases can indicate the feasibility of the analysis of a sentence at the level of phrases. Moreover, it can imply the necessity of taking into account the availability of coreferent relations (except for cataphora).

- The highest impact of metric *SemCoh* may indicate the feasibility to take into account the connection between all text elements notwithstanding its position within a text.

**Відомості про авторів**

*Крамов Артем Андрійович*

Науковий ступінь: –.

Вчене звання: –.

Посада: аспірант кафедри комп'ютерної інженерії факультету радіофізики, електроніки та комп'ютерних систем Київського національного університету імені Тараса Шевченка.

Установа: Київський національний університет імені Тараса Шевченка.

Домашня адреса: 38800, смт. Чутове, Полтавська область, вул. Полтавський Шлях, 87.

Службова адреса: 03022, Київ, проспект Академіка Глушкова, 4Г.

Телефон: +38 (050) 149-31-32

E-mail: artemkramov@gmail.com (для роботи з редактором)